\documentclass[journal, twocolumn]{IEEEtran}
\IEEEoverridecommandlockouts
\usepackage{graphicx}
\usepackage{xcolor}
\usepackage[skip=1pt]{subcaption}
\usepackage{amsmath}
\usepackage{gensymb}
\usepackage{amssymb}
\usepackage{algorithm}
\usepackage{algorithmicx}
\usepackage[noend]{algpseudocode}
\usepackage{booktabs}
\usepackage{titlesec}
\usepackage{listings}
\usepackage{enumitem}
\usepackage{multirow}
\usepackage{placeins}
\usepackage{rotating}
\usepackage{pifont}
\usepackage{array}
\usepackage{mdframed}
\usepackage{lipsum}
\usepackage[utf8]{inputenc}
\usepackage[many]{tcolorbox}
\usepackage{varwidth}
\usepackage{colortbl}
\usepackage{array}
\usepackage[colorlinks=true, allcolors=blue]{hyperref}
\usepackage{cleveref}
\usepackage{makecell}

\usepackage[font=scriptsize]{caption}
\title{
\fontsize{13.75}{13.75} \selectfont
PathBench: A Benchmarking Platform for Classical and Learned Path Planning Algorithms \vspace{-2 mm}}

\author{
Alexandru-Iosif Toma$^{\dagger}$, 
Hao-Ya Hsueh$^{\star}$,  
Hussein Ali Jaafar$^{\star}$, 
Riku Murai$^{\dagger}$, 
Paul H.J. Kelly$^{\dagger}$,  
Sajad Saeedi$^{\star}$
\vspace{-7 mm}
\thanks{$^{\dagger}$Imperial College London\quad\quad $^{\star}$Ryerson University}%
}

\begin{document}
\maketitle

%===============================================================================

%%%%%%%%%%%%%%%%%%%%%%%%%%%%%%%%%%%%%%%%%%%%%%%%%%%%%%%%%%%%%
%%%%%%% Abstract
%%%%%%%%%%%%%%%%%%%%%%%%%%%%%%%%%%%%%%%%%%%%%%%%%%%%%%%%%%%%%

%\thispagestyle{empty}
%\pagestyle{empty}

\begin{abstract}

Path planning is a key component in mobile robotics. A wide range of path planning algorithms exist, but few attempts have been made to benchmark the algorithms holistically or unify their interface. Moreover, with the recent advances in deep neural networks, there is an urgent need to facilitate the development and benchmarking of such learning-based planning algorithms. This paper presents PathBench, a platform for developing, visualizing, training, testing, and benchmarking of existing and future, classical and learned 2D and 3D path planning algorithms, while offering support for Robot Operating System (ROS). Many existing path planning algorithms are supported; e.g. A*, wavefront, rapidly-exploring random tree,  value iteration networks, gated path planning networks; and integrating new algorithms is easy and clearly specified. We demonstrate the benchmarking capability of PathBench by comparing implemented classical and learned algorithms for metrics, such as path length, success rate, computational time and path deviation. These evaluations are done on built-in PathBench maps and external path planning environments from video games and real world databases. PathBench is open source\footnote{\href{https://sites.google.com/view/PathBench}{https://sites.google.com/view/PathBench}}.%\footnote{\href{https://github.com/perfectly-balanced/PathBench}{https://github.com/perfectly-balanced/PathBench}}.
\end{abstract}

\hspace{-2 mm}

%===============================================================================

%%%%%%%%%%%%%%%%%%%%%%%%%%%%%%%%%%%%%%%%%%%%%%%%%%%%%%%%%%%%%
%%%%%%% Introduction
%%%%%%%%%%%%%%%%%%%%%%%%%%%%%%%%%%%%%%%%%%%%%%%%%%%%%%%%%%%%%
\section{Introduction}
Path planning is an optimization problem with one or multiple objectives that aims to find a desirable path between two poses. Autonomous robots rely on various path planning algorithms to meet specific performance metrics \cite{gonzalez2016review}. In robotics and motion planning, benchmarking and comparison between algorithms is key to the experimental evaluation of newly proposed algorithms. Papers often report performance scores such as length or curvature of the path, but also increasingly complex metrics, such as execution time and memory consumption. However, as the diversity of the algorithms is expanding, especially with the advances in the areas of machine learning (ML), it becomes a challenging issue to benchmark algorithms efficiently. To address this issue, we present a unified framework that supports benchmarking and development of classical and learned planning algorithms. 

\begin{figure}[t]
  \centering
%  \begin{subfigure}[b]{0.2\linewidth}
%    \includegraphics[width=\linewidth]{images/screenshot_43.png}
%     \caption{\scriptsize Wavefront}
%  \end{subfigure}
  %\hfill
    \begin{subfigure}[b]{0.3\linewidth}
    \includegraphics[width=\linewidth]{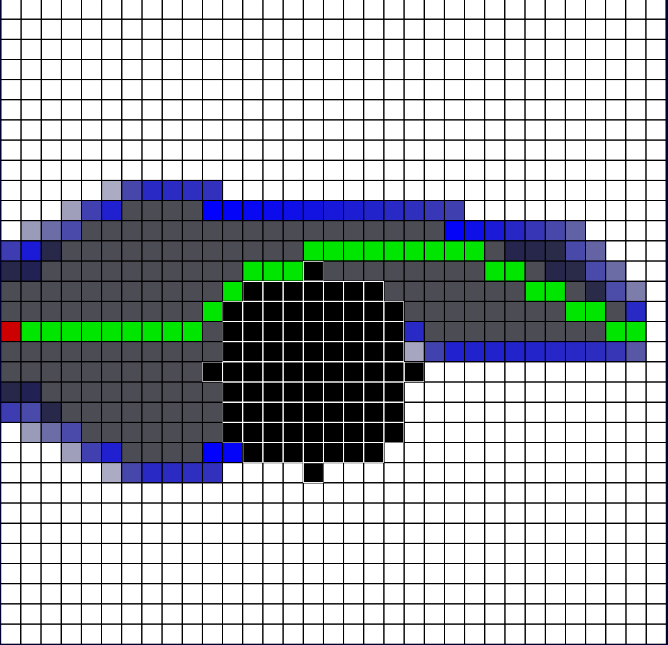}
     \caption{\scriptsize A*}
  \end{subfigure}
  \hfill
  \begin{subfigure}[b]{0.3\linewidth}
    \includegraphics[width=\linewidth]{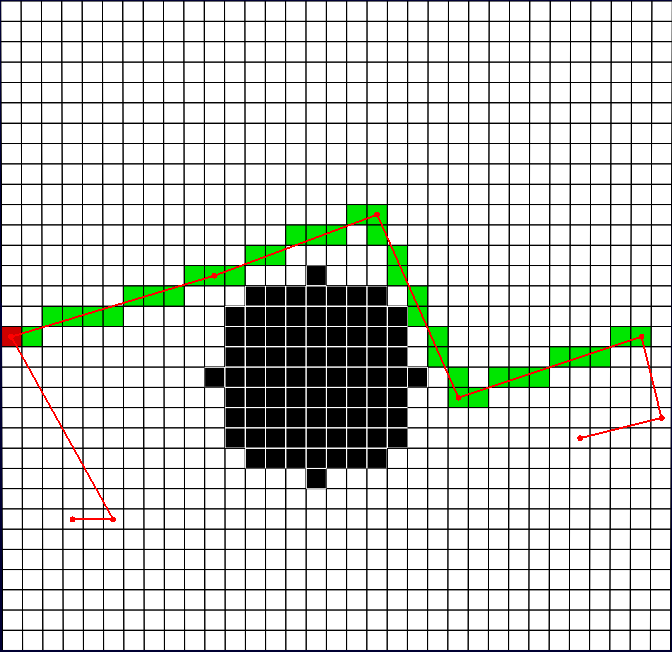}
    \caption{\scriptsize RRT-Connect}
  \end{subfigure}
  \hfill
  \begin{subfigure}[b]{0.3\linewidth}
    \includegraphics[width=\linewidth]{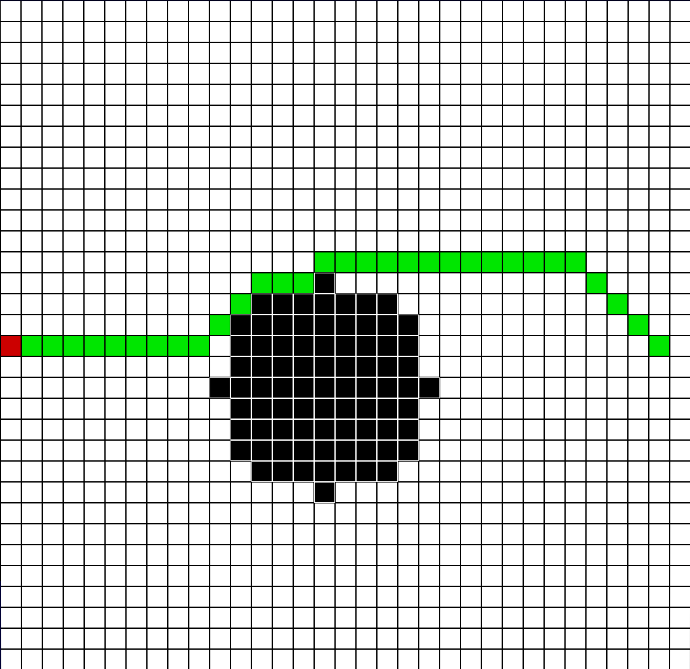}
    \caption{\scriptsize GPPN }
  \end{subfigure}
  \begin{subfigure}[b]{0.3\linewidth}
    \includegraphics[width=\linewidth]{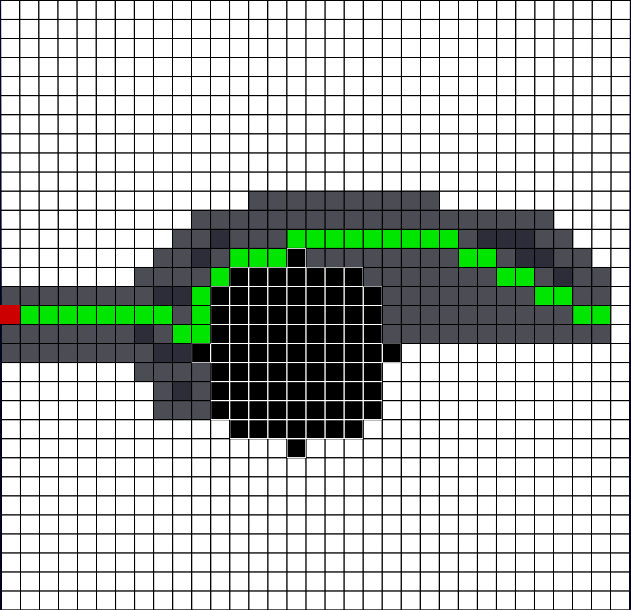}
     \caption{\scriptsize {WPN}}
  \end{subfigure}
  \hfill
  \begin{subfigure}[b]{0.3\linewidth}
    \includegraphics[width=\linewidth]{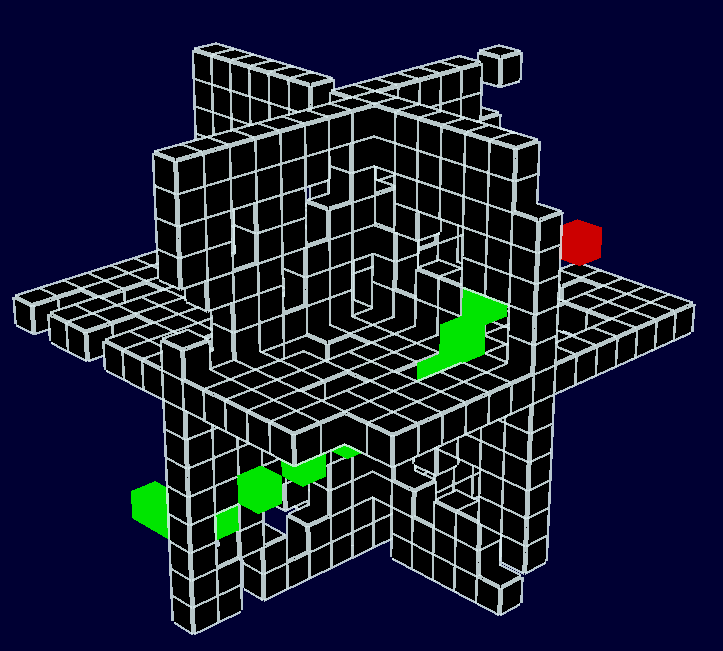}
     \caption{\scriptsize A* }
     \vspace{0.5pt}
  \end{subfigure}
  \hfill
  \begin{subfigure}[b]{0.31\linewidth}
    \includegraphics[width=\linewidth]{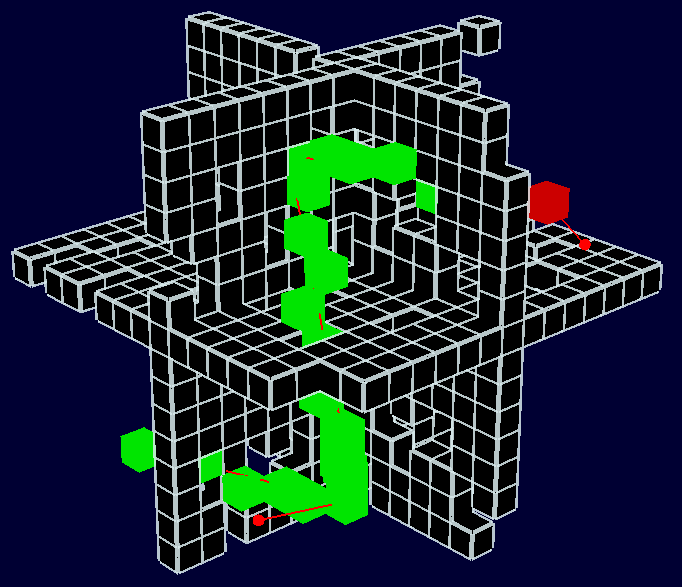}
     \caption{\scriptsize RRT-Connect}
  \end{subfigure}
 % \hfill
 % \begin{subfigure}[b]{0.2\linewidth}
 %   \includegraphics[width=\linewidth]{images/2dwavefront.png}
 %   \caption{\scriptsize Wavefront}
 % \end{subfigure}
  \vspace{1 mm}
  \caption{The results of different classical and learned planners in PathBench. The red entity is the agent, light green entities are traces to the goal, black/light gray entities are obstacles and everything else is custom display information (e.g. in (a) and (d), dark gray represents the search space)}
  \label{fig: sim}
      \vspace{- 3mm}
\end{figure}

In this paper we introduce PathBench, a motion planning platform that can be used to develop, assess, compare, and visualize the performance and behaviour of path planning algorithms, Fig. \ref{fig: sim}. 
PathBench has three key features. 
(1) It supports both 2D and 3D classical and learned algorithms. Existing machine learning based algorithms, such as value iteration networks 
(VIN) \cite{tamar2016value}, 
gated path planning networks (GPPN) \cite{lee2018gated}, 
motion planning networks (MPNet) \cite{qureshi2019motion}, as well as Online LSTM~\cite{nicola2018lstm}, and CAE-LSTM~\cite{inoue2019robot} methods, are incorporated into PathBench. PathBench has a structured environment to facilitate easy development and integration of new classical and ML-based algorithms. (2) PathBench's benchmarking features allow evaluation against the suites of added path planning algorithms, both classical algorithms and machine-learned models, with standardized metrics and environments. (3) PathBench provides a ROS (Robot Operating System) real-time extension for interacting with a real-world robot. Examples are provide for these features. 

%%%%%%%%%%%%%%%%%%%%%%%%%%%%%%%%%%%%%%%%%%%%%%%%%%%%%%%%%%%%%
%%%%%%% Related Work
%%%%%%%%%%%%%%%%%%%%%%%%%%%%%%%%%%%%%%%%%%%%%%%%%%%%%%%%%%%%%
\section{Related Work}
In this section, classical and learned planning algorithms, and existing benchmarking frameworks are reviewed briefly.

\subsection{Classical and Learned Planning Algorithms}
There are four general categories for planning algorithms: graph search, sampling-based, sensory-based, and numerical optimization \cite{gonzalez2016review}. Graph search algorithms typically operate on grid and lattice maps, representing discretized state spaces. Popular examples of such algorithms are, 
Dijkstra 
\cite{choset2005principles},  %zhang2014multiple
A* 
\cite{duchovn2014path}, 
%choset2005principles, 
%zhang2014multiple, 
%5937169
and wavefront \cite{luo2014effective}. Rapidly-exploring random tree (RRT) \cite{lavalle1998rapidly} and probabilistic roadmap (PRM) \cite{kavraki1994probabilistic}, are examples of sampling-based planning algorithms. These algorithms randomly sample the configuration space or the state space to create a path. In high-dimension planning problems, the sampling-based algorithms work efficiently in comparison with graph-based methods in term of computational resources, at the cost of being non-optimal. 
In contrast to the sampling and graph based methods, the sensor-based planning algorithms only plan for the view~\cite{sensor1997, Paull_TMech_2013}, i.e local maps. Examples of such algorithms include Bug1 and Bug2~\cite{choset2005principles, rajko2001pursuit}. The fourth category, the numerical optimization  planners, operate by optimizing a cost function composed of one or multiple terms. The cost functions could include various constraints such as kinematics \cite{ziegler2014making} and the smoothness of the trajectory \cite{dolgov2010path}. The recent deep learning methods, such as CoMPNet~\cite{CoMPNet} and VIN~\cite{tamar2016value}, are also considered numerical optimization planners.

%\subsection{Learned Planning Algorithms}
The availability of large-scale data and parallel processing systems has shifted the attention of the researcher to learning-based planning algorithms~\cite{inoue2019robot,Chen2016Humanoids,gupta2017cognitive}. Some of the planning algorithms aim at improving specific parts of the classical algorithms. For instance, Qureshi {\it et al.} perform sampling on particular regions of the configuration space as opposed to the whole space~\cite{qureshi2018deeply}. Using a similar approach, Chamzas {\it et al.} reduce the computational complexity of the classical planning algorithms \cite{chamzas2019using}.
Other planning algorithms generate full paths via neural networks. Motion planning networks (MPNet), generate a path from start to goal via a trained neural network which takes a point cloud map as an input~\cite{qureshi2018motion}. The recent version of MPNet accounts for the kinematics constraint as well \cite{CoMPNet}, \cite{qureshi2019motion}. Various architectures and machine learning methods are being used in planning including: 
recurrent neural networks (RNN) in OracleNet~\cite{bency2019neural}, 
3D supervised imitation learning in (TDPP-Net) \cite{TDPPNet}, 
unsupervised Generative Adversarial Networks (GANs) in \cite{mohammadi2018path} and \cite{choi2020pathgan}, 
and reinforcement learning (RL) strategies in value iteration networks (VIN)~\cite{tamar2016value}, gated path planning networks (GPPN) \cite{lee2018gated}, universal planning networks (UPN) \cite{srinivas2018universal}  \cite{Levine2013}, guided policy search (GPS) \cite{Levine2013}, and learning-from demonstration (LfD)~\cite{Abbeel2010}.
With the continual advancement in machine and deep learning techniques and hardware capabilities, increased development of new learning based path planning algorithms can be foreseen.

%%%%%%%%%%%%%%%%%%%%%%%%%%%%%%%%%%%%%%%%%%%%%%%%%%%%%%%%%%%%%
%%%%%%% Simulation and benchmarking platforms
%%%%%%%%%%%%%%%%%%%%%%%%%%%%%%%%%%%%%%%%%%%%%%%%%%%%%%%%%%%%%
\subsection{Simulation and Benchmarking Platforms}
{Benchmarking} of path planning algorithms is the scientific approach of evaluation in the robotics community. 
Currently, there are a variety of standardized libraries %which contain at least two of the mentioned sections (\textbf{Simulator} and \textbf{Analyzer})
relevant to path planning, such as \textit{ROS} \cite{Quigley09}, \textit{OpenRAVE} \cite{diankov2008openrave}, \textit{OMPL} \cite{sucan2012the_open_motion_planning_library},  \textit{MoveIt} \cite{moveit} (which has benchmarking capabilities \cite{moll2015benchmarking}), \textit{SBPL} \cite{plaku2007oops}, and \textit{OOPS\textsubscript{MP}} \cite{plaku2007oops}. These will briefly be summarized, before comparing to our platform.

%Currently, there are a variety of standardised libraries which contain at least two of the mentioned sections (\textbf{Simulator} and \textbf{Analyzer}) such as: \textit{ROS} \cite{Quigley09}, \textit{OMPL} \cite{sucan2012the_open_motion_planning_library}, \textit{MoveIt} \cite{moveit} (has benchmarking capabilities \cite{moll2015benchmarking}).

{\it ROS.} The Robot Operating System (ROS) is a middleware which contains various planning algorithms and simulation environments (including 2D and 3D) for different types of robots: ground robots with different degrees of freedom constraints, flying robots (drones), and manipulator robots. % (arm robots which interact with different objects). 
ROS is the standard in robotics for simulation and development.

{\it OpenRAVE.} The Open Robotics and Animation Virtual Environment (OpenRAVE), is an open-source cross-platform software architecture, targeted for real-world robots, which includes 3D simulation, visualization, planning, scripting, and control. Compared to ROS, it is focused on autonomous motion planning and high-level scripting rather than low-level control and message protocols.

{\it OMPL.} The Open Motion Planning Library (OMPL) is a standalone library which focuses on motion planning exclusively. It is more lightweight than ROS and has reduced capabilities (there is no collision detection). The library of available path planners is limited to sampling-based planners such as RRT or PRM, but there is a variety of optimized implementations for each type of planner. OMPL has a benchmarking extension called: planner arena, where a community-contributed database of benchmarking data allows for visualization of algorithm comparison.

{\it SBPL.} The Search-Based Planning Library (SBPL) is a small library of graph search implementations with 2D and 3D environments, but no benchmarking capability.

{\it MoveIt.} Combines both ROS and OMPL to create a high-level implementation for cleaner and faster development of new algorithms. It has more capabilities than ROS and OMPL and includes custom benchmarking techniques \cite{moll2015benchmarking}.

{\it OOPS\textsubscript{MP}}. The Online, Open-source, Programming System for Motion Planning (OOPS\textsubscript{MP}) is an online platform for comparing sampling-based motion planners on a common set of 2D and 3D maps, also providing implementations of common algorithms, with analysis and visualization tools for benchmarking.

{\it PathBench.} Our implementation offers a more abstract overview of the environment, naturally allowing for graph-based, sampling-based, and machine learning based approaches to path planning. It supports 2D and 3D environments. PathBench includes not only a simulation environment 
%({simulator}) 
and benchmarking techniques,
%(
%\textbf
%({analyzer})
but also a generator for creating synthetic datasets for ML applications 
%\textbf
%({generator}) 
and a ML training pipeline 
%\textbf
%({Trainer}) 
for generic ML models.

%%%%%%%%%%%%%%%%%%%%%%%%%%%%%%%%%%%%%%%%%%%%%%%%%%%%%%%%%%%%%
%%%%%%% Table
%%%%%%%%%%%%%%%%%%%%%%%%%%%%%%%%%%%%%%%%%%%%%%%%%%%%%%%%%%%%%
\begin{table}[t]
    \footnotesize
    \center
    \caption{Platform capabilities comparison. PathBench supports benchmarking of classical and learned planning algorithms. %H - High, M - Moderate, R - Reduced. ECI is short for Environment Complexity and Interaction
    }
    \vspace{1mm}
    \begin{tabular}{|c| c c | c c c |}% c c c |}
         \hline
         {\scriptsize \rotatebox[origin=c]{0}{Platform}} & {\scriptsize \rotatebox[origin=c]{60}{Visualization}} & {\scriptsize \rotatebox[origin=c]{60}{Benchmarking}} & {\scriptsize \rotatebox[origin=c]{60}{Sample-Based}} &{\scriptsize \rotatebox[origin=c]{60}{Graph-Based}} &  {\scriptsize \rotatebox[origin=c]{60}{ML-Based}} \\%&  {\scriptsize \rotatebox[origin=c]{90}{Efficiency}} & {\scriptsize \rotatebox[origin=c]{90}{Variety}} & {\scriptsize \rotatebox[origin=c]{90}{ECI}} \\
         \hline
         {ROS} & \textcolor{green}{\checkmark} & \textcolor{red}{$\times$} & \textcolor{green}{\checkmark} & \textcolor{green}{\checkmark} & \textcolor{red}{$\times$} \\% & {R} & {H} & {H} \\
         \hline
         {OpenRAVE} & \textcolor{green}{\checkmark} & \textcolor{red}{$\times$} & \textcolor{green}{\checkmark} & \textcolor{red}{$\times$} & \textcolor{red}{$\times$}  \\% {M} & {H} & {H} \\
         \hline
         {OMPL} & \textcolor{green}{\checkmark} & \textcolor{green}{\checkmark} & \textcolor{green}{\checkmark} & \textcolor{red}{$\times$} & \textcolor{red}{$\times$} \\% & {H} & {R} & {R} \\
         \hline
         {MoveIt} & \textcolor{green}{\checkmark} & \textcolor{green}{\checkmark} & \textcolor{green}{\checkmark} & \textcolor{red}{$\times$} & \textcolor{red}{$\times$} \\% & {M} & {H} & {H} \\
         \hline
         {SBPL} & \textcolor{green}{\checkmark} & \textcolor{red}{$\times$} & \textcolor{red}{$\times$} & \textcolor{green}{\checkmark} & \textcolor{red}{$\times$}  \\% {R} & {R} & {M} \\
         \hline
         {OOPS\textsubscript{MP}} & \textcolor{green}{\checkmark} & \textcolor{green}{\checkmark} & \textcolor{green}{\checkmark} & \textcolor{red}{$\times$} & \textcolor{red}{$\times$} \\% & {M} & {M} & {M} \\
         \hline
         \textbf{PathBench} & \textcolor{green}{\checkmark} & \textcolor{green}{\checkmark} & \textcolor{green}{\checkmark} & \textcolor{green}{\checkmark} & \textcolor{green}{\checkmark} \\% & {H} & {M} & {R} \\
         \hline
    \end{tabular}
    \label{tab: plat_comparison}
        \vspace{-5 mm}
\end{table}

Further to these standard libraries, other works have sought to address the issue of benchmarking for motion planning. Sturtevan provides a standard test set of maps, and suggest standardized metrics for grid based planning for gaming environments~\cite{sturtevant2012benchmarks}. These maps have been ported into PathBench and are discussed further in Sec.~\ref{sec:maps}. Althoff {\it et al.} provide a collection of composable benchmarks for motion planning of cars on roads, allowing reproducible results on problems~\cite{althoff2017commonroad}. %[ref] do something else. 

The main advantages of PathBench over existing standardized libraries are its native support of machine learning path planning algorithms, as well as its simple, lightweight, and extensible design, allowing fast prototyping for a research environment. It provides a standardized set of maps and metrics, so that benchmarking of new and existing algorithms can be performed quickly. Moreover, we provide a clean API interface for the algorithms which makes them portable to the standardized libraries. Also, we provide a {ROS} real-time extension which converts the internal map move actions into network messages (velocity control commands) using the 
{ROS} APIs. See Table~\ref{tab: plat_comparison} for platform comparison.

%%%%%%%%%%%%%%%%%%%%%%%%%%%%%%%%%%%%%%%%%%%%%%%%%%%%%%%%%%%%%
%%%%%%% PathBench Platform
%%%%%%%%%%%%%%%%%%%%%%%%%%%%%%%%%%%%%%%%%%%%%%%%%%%%%%%%%%%%%
\section{PathBench Platform}

\begin{figure}[t]
    \centering
    %\hspace{-10 mm}
    \includegraphics[scale=0.4]{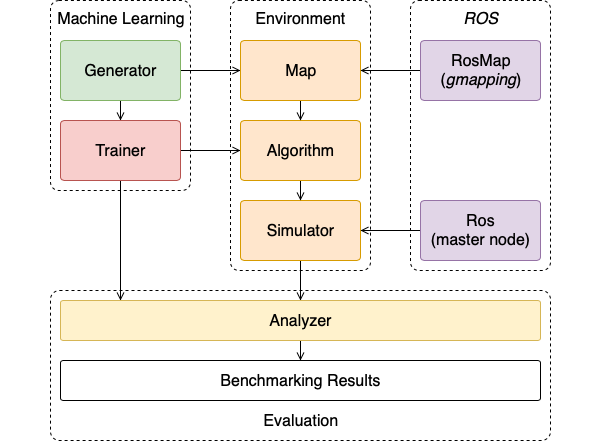}
     \vspace{1mm}
    \caption{PathBench structure overview. Arrows represent information flow/usage ($A \xleftarrow{gets/uses} B$). The machine learning section is responsible for training dataset generation and model training. The Environment section controls the interaction between the agent and the map, and supplies graphical visualization. The \textit{ROS} section provides support for real-time interaction with a real physical robot. The Evaluation section provides benchmarking methods for algorithm assessment. For a detailed architecture, please refer to the website of PathBench.}
    \label{fig: sim_platform}
        \vspace{-5 mm}
\end{figure}

An overview of the architecture of PathBench is shown in Fig. \ref{fig: sim_platform}. PathBench is composed of four main components: \emph{Simulator}, \emph{Generator}, \emph{Trainer}, and \emph{Analyzer} joined by the \emph{infrastructure} section. 

%\textbf{Infrastructure.} This component 
The infrastructure is responsible for linking all other components and provides general service libraries and utilities.
%
%\textbf{Simulator.} This section 
The simulator is responsible for environment interactions and algorithm visualization. It provides custom collision detection systems and a graphics framework for rendering the internal state of the algorithms.
%
%\textbf{Generator.} This section 
The generator is responsible for generating and labelling the training data used to train the ML models.
%
%\textbf{Trainer.} This section 
The trainer is a class wrapper over the third party machine learning libraries. It provides a generic training pipeline based on the holdout method and standardized access to the training data.
%
%\textbf{Analyzer.} The final section 
Finally, the analyzer manages the statistical measures used in the practical assessment of the algorithms. Custom metrics can be defined, as well as graphical displays for visual comparisons. 
PathBench has been written in Python, and uses PyTorch \cite{paszke2017automatic} for ML.

%%%%%%%%%%%%%%%%%%%%%%%%%%%%%%%%%%%%%%%%%%%%%%%%%%%%%%%%%%%%%
%%%%%%% Figure
%%%%%%%%%%%%%%%%%%%%%%%%%%%%%%%%%%%%%%%%%%%%%%%%%%%%%%%%%%%%%

%%%%%%%%%%%%%%%%%%%%%%%%%%%%%%%%%%%%%%%%%%%%%%%%%%%%%%%%%%%%%
%%%%%%% Simulator
%%%%%%%%%%%%%%%%%%%%%%%%%%%%%%%%%%%%%%%%%%%%%%%%%%%%%%%%%%%%%
\subsection{Simulator}

\begin{figure}[t]
    \centering
    %\hspace{-10 mm}
    \includegraphics[scale=0.2]{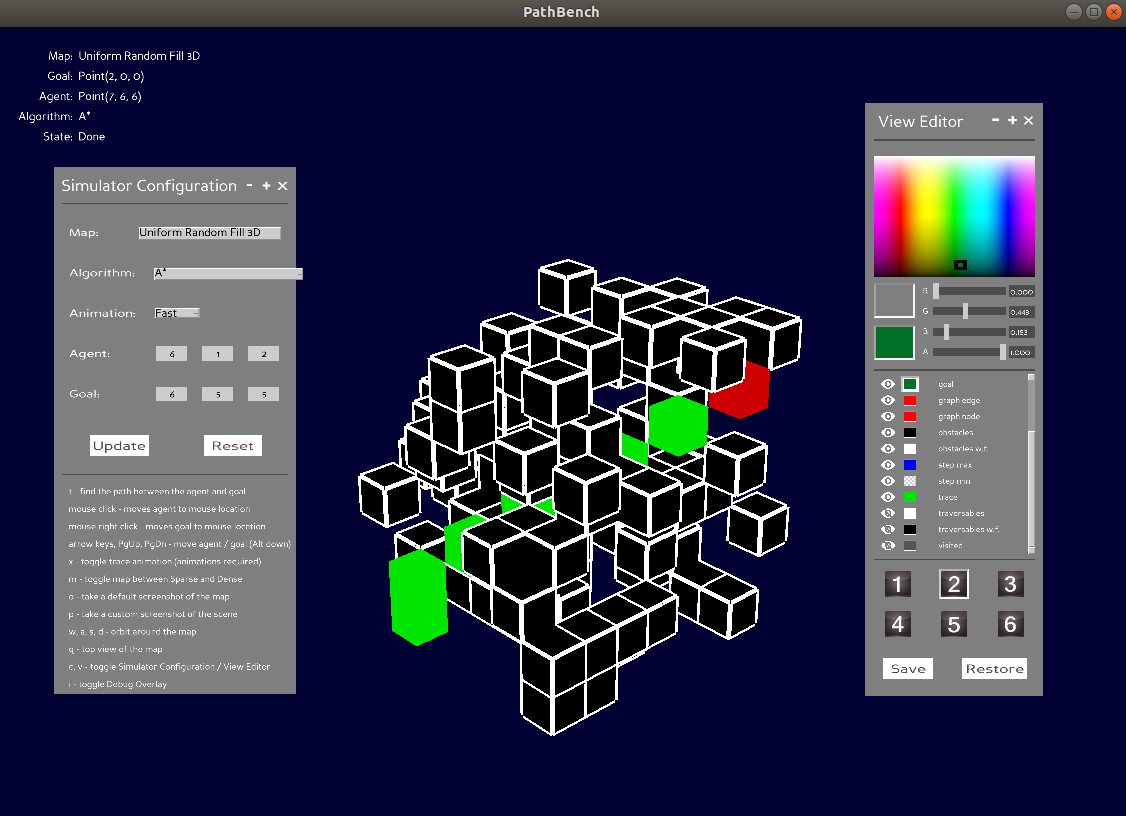}
     \vspace{1mm}
    \caption{ The GUI of the simulator is where configuration of the path planners, environments and customization of the visualization are made interactively. Simulator configuration window is used to set up path planning sessions. View editor allows for adjustment of visualized environment appearance.}
    \label{fig: simulator3d}
        \vspace{-5 mm}
\end{figure}

The {simulator} is both a visualizer and an engine for developing algorithms (Fig. \ref{fig: sim}). It supports animations and custom map display components which render the {algorithm}'s internal data. Simulator has a {map} that contains different entities such as the {agent}, {goal} and {obstacle}s, and provides a clean interface that defines the movement and interaction between them. Therefore, a {map} can be extended to support various environments; however, each map has to implement its own physics engine or use a third party one (e.g. the 
\textit{pymunk} physics engine 
%\cite{pymunk} 
or \textit{OpenAI Gym} 
%\cite{OpenAI_Gym}
). The current implementation supports three types of 2D/3D maps: {DenseMap}, {SparseMap} and {RosMap}, corresponding to static grid map, point cloud map, and grid map with live updates, respectively. Additionally, the {simulator} provides animations that are achieved through key frames and synchronization primitives. The graphical framework used for the visualization of planners and GUI is Panda3D \cite{panda}. Simulator configurations and visualization customizations can be directly controlled within the Panda3D GUI, see Fig.~\ref{fig: simulator3d}.

%%%%%%%%%%%%%%%%%%%%%%%%%%%%%%%%%%%%%%%%%%%%%%%%%%%%%%%%%%%%%
%%%%%%% Generator
%%%%%%%%%%%%%%%%%%%%%%%%%%%%%%%%%%%%%%%%%%%%%%%%%%%%%%%%%%%%%
\subsection{Generator} \label{sec: generator}

The {generator} can execute four actions: (1) map generation, (2) map labelling, (3) map augmentation, and (3) map modification, each explained briefly below.

{\it 1) Generation.} The generation procedure accepts as input, different hyper-parameters such as the type of generated maps, number of generated maps, number of dimensions, obstacle fill rate range, number of obstacle range, minimum room size range and maximum room size range. 
Currently, the generator can produce four types of maps: uniform random fill map, 
block map, house map and point cloud map,
(See Fig. \ref{fig:3maps}) 
and it can be extended to support different synthetic maps such as mazes and cave generation using cellular automata. All generated maps are placed into a directory in both .pickle and JSON formats.

{\it 2) Labelling.} The labelling procedure takes a map and converts it into training data by picking only the specified features and labels. 
%The training data is then saved as a \texttt{.pickle} file with name format as \texttt{training\_\{atlas name\}\_\{number of samples\}}. The structure of the training data is based on normal \textit{python} objects (\texttt{List[Dict[str, Any]]}) for quick inspection and analysis. 
Features/labels %are picked by using the {MapProcessing} component. These data 
include agent and goal positions, global map, local view, valid moves, etc.  
%(See Table \ref{tab: gen_label_list} for feature reference). 
A* is used as ground truth for feature/label generation. All features/labels can be saved as a variable sequence (needed for LSTM) or single global input (needed for auto-encoder).

{\it 3) Augmentation.} The augmentation procedure takes an existing training data file and augments it with the specified extra features and labels. It is used to remove the need for re-generating a whole training set.

{\it 4) Modification.} A custom lambda function which takes as input a {map} and returns another {map} and can be defined to modify the underlining structure of the map (e.g. modify the agent position, the goal position, create doors, etc.).

%%%%%%%%%%%%%%%%%%%%%%%%%%%%%%%%%%%%%%%%%%%%%%%%%%%%%%%%%%%%%
%%%%%%% Figure
%%%%%%%%%%%%%%%%%%%%%%%%%%%%%%%%%%%%%%%%%%%%%%%%%%%%%%%%%%%%%
%In the rest of the section, we describe the PathBench platform and its implementation.
%An overview of the PathBench platform design is given in Fig. \ref{fig:sim_architecture}. 

%%%%%%%%%%%%%%%%%%%%%%%%%%%%%%%%%%%%%%%%%%%%%%%%%%%%%%%%%%%%%
%%%%%%% Trainer
%%%%%%%%%%%%%%%%%%%%%%%%%%%%%%%%%%%%%%%%%%%%%%%%%%%%%%%%%%%%%
\subsection{Trainer}

The training pipeline is composed of: (1) data pre-processing, (2) data splitting, (3) training, (4) evaluation, (5) results display, and (6) pipeline end, explained below briefly.

{\it 1) Data Pre-processing.} Data is loaded from the specified training sets, and only the features and labels used throughout the model are picked from the training set and converted to a PyTorch {dataset}. In total, there can be four datasets: one feature sequence, one single feature tensor, one label sequence, and one single label tensor. Sequential data is wrapped into a {PackedDataset} which sorts the input in reverse order of its sequence length (max length first, min length last). 

{\it 2) Data Splitting.} The pre-processed data is shuffled and split into three categories: training, validation and testing (usually 60\%, 20\%, 20\%) according to the holdout method. The {CombinedSubsets} object is used to couple the feature dataset and label dataset of the same category into a single dataset. Then, all data is wrapped into its {DataLoader} object with the same batch size as the training configuration (usually 50).

{\it 3) Training.} The training process puts the model into training mode and takes the training {DataLoader} and validation {DataLoader} and feeds them through the model $n$ times, where $n$ is the number of specified epochs. The training mode allows the gradients to be updated and at each new epoch, the optimizer sets all gradients to 0. Each model has to extend a special \texttt{batch\_start} hook function which is called on each new batch. The \texttt{batch\_start} function is responsible for passing the data through the network and returning the loss result. The trainer takes the loss result and applies a backward pass by calling the \texttt{.backward()} method from the loss. Afterwards, the optimizer is stepped, and the weights of the model are updated. The statistics, such as the loss over time, for the training and validation sets are logged by two {EvaluationResults} objects (one for training and one for validation) which are returned to the pipeline. The {EvaluationResults} class contains several hook functions which are called through the training process at their appropriate times: \texttt{start}, \texttt{epoch\_start}, \texttt{epoch\_finish}, \texttt{batch\_start}, \texttt{batch\_finish}, \texttt{finish}. At each epoch end, the {EvaluationResults} object prints the latest results.

{\it 4) Evaluation.} The evaluation process puts the model into evaluation mode and has a similar structure to the training process. The evaluation mode does not allow gradients to update. The testing dataset is passed only once through the model and an {EvaluationResults} object containing the final model statistics is returned to the pipeline.

{\it 5) Results Display.} This procedure displays the final results from the three {EvaluationResults} objects (training, validation, testing) and final statistics such as the model loss are printed. The training and validation loss logs are displayed as a \textit{matplotlib} \cite{Hunter:2007} figure. This method can be easily extended to provide more insight into the network architecture (e.g. the covolutional autoencoder {(CAE)} model displays a plot which contains the original image, the reconstructed version, the latent space snapshot and the resulting feature maps).

%\subsubsection*{{Pipeline End}} 
{\it 6) Pipeline End.} At the end, the model is saved by serialising the model \texttt{.state\_dict()}, model configuration, plots from results display process, and full printing log. % into a {ModelSudir} under {ModelDir}. %The save name is formated according to the following convention: \texttt{\{config save\_name\}\_\{config training\_data\}\_model}.

%%%%%%%%%%%%%%%%%%%%%%%%%%%%%%%%%%%%%%%%%%%%%%%%%%%%%%%%%%%%%
%%%%%%% Analyzer
%%%%%%%%%%%%%%%%%%%%%%%%%%%%%%%%%%%%%%%%%%%%%%%%%%%%%%%%%%%%%
\subsection{Analyzer}

The {analyzer} is used to assess and compare the performance of the path planners. This is achieved by making use of the {BasicTesting} component. When a new session is run through the {AlgorithmRunner}, statistical measures depending on the type of testing can be collected by attaching a {BasicTesting} component.
%(See Table \ref{tab: a_testing_tab}). 
The {BasicTesting} component is also linked to the simulator to enable visualization testing. The key frame feature and synchronization variable are tied to the {BasicTesting} component, which allows the user to enhance each key frame and define custom behaviour. Each {algorithm} instance can create debugging views called {MapDisplay}s which can render custom information on the screen such as the the internal state of the {algortihm} (e.g. search space, total fringe, graph, map and its entities etc).% (See Table \ref{tab: map_dsiplays}).

%Instead of manually running a \textbf{Simulator} instance to assess an \textbf{Algorithm}, the \textbf{analyzer} has an extensive algorithmic analysis procedure split into two parts: simple analysis and complex analysis. We also provide a training dataset analysis routine for inspecting the generated maps.

In addition to manually running a {simulator} instance to assess an {algorithm}, the {analyzer} supports the following analysis procedures:

\begin{itemize}
\addtolength{\itemindent}{-.4cm}
\item {Simple Analysis.} $n$ map samples are picked from each generated map type, and $m$ algorithms are assessed on them. The results are averaged and printed. Barplots and violinplots, for metrics discussed in Sec.~\ref{sec:performance metrics}, are generated with results from Simple Analysis.
\item {Complex Analysis.} $n$ maps are selected (generated or hand-made), and $m$ algorithms are run on each map $x$ (usually 50) times with random agent and goal positions.  %As in the simple analysis stage, 
In the end, all $n \times x$ results are averaged and reported. Similarly to Simple Analysis, barplots and violinplots for selected metrics can be generated with results.
\item {Training Dataset Analysis.} A training set analyzer procedure is provided to inspect the training datasets by using the basic metrics 
%defined in Table \ref{tab: a_testing_tab} 
(e.g. Euclidean distance, success rate, map obstacle ratio, search space, total fringe, steps, etc. See the website of the project for all metrics and statistics).   
\end{itemize}

\begin{figure}[t]
  \begin{minipage}[b]{0.32\columnwidth}
  \centering
    \includegraphics[width=.99\columnwidth]{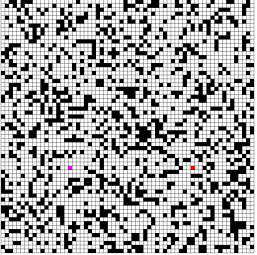}\\
    \includegraphics[width=.95\columnwidth]{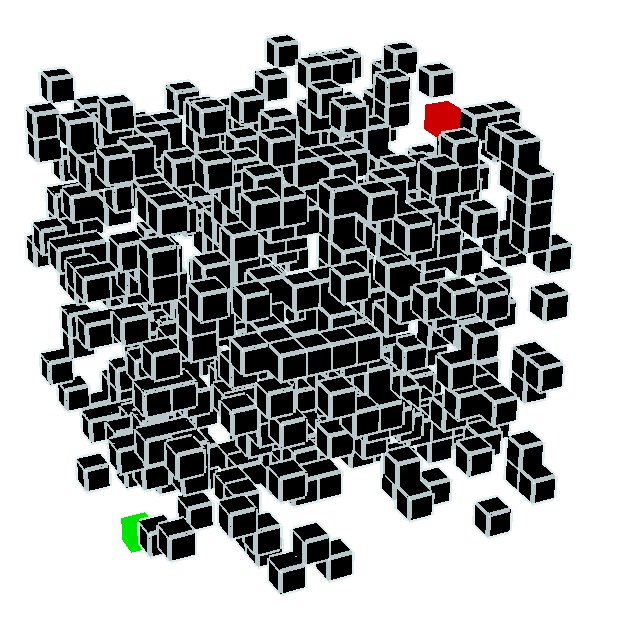} \\
    (a) 
  \end{minipage}
  \begin{minipage}[b]{0.32\columnwidth}
  \centering
    \includegraphics[width=.99\columnwidth]{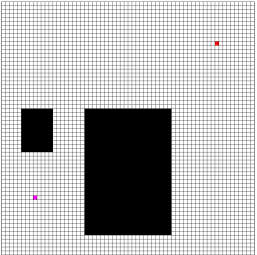}\\
    \includegraphics[width=.99\columnwidth]{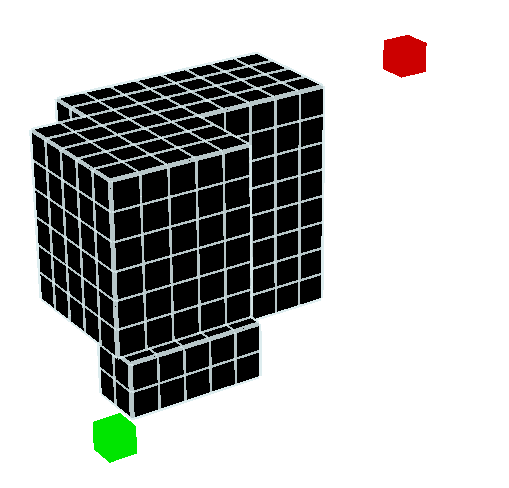} \\
    (b)
  \end{minipage}
  \begin{minipage}[b]{0.32\columnwidth}
  \centering
    \includegraphics[width=0.99\columnwidth]{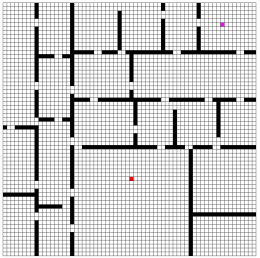}\\
    \includegraphics[width=.99\columnwidth]{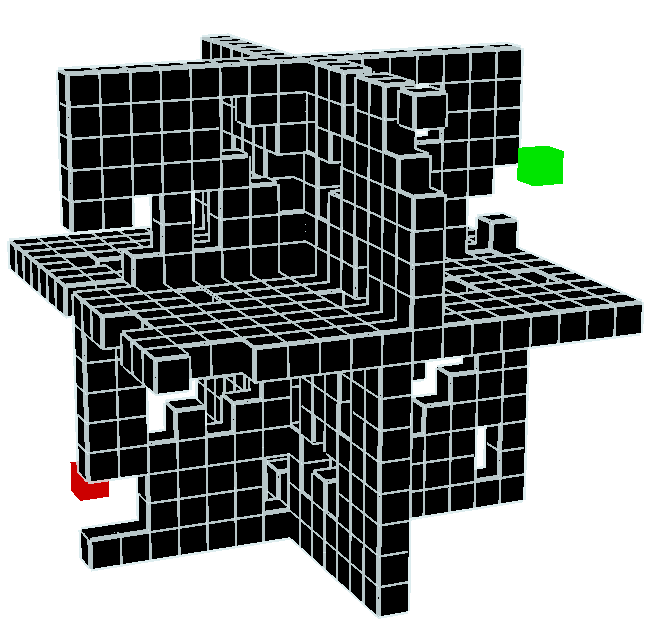} \\
    (c)
  \end{minipage}
  \vspace{1mm}
\caption{The 3 types of generatable grid maps are: (a) Uniform random fill ($64\times64$ 2D and $16\times16$ 3D dimensions, [0.1, 0.3] obstacle fill rate range), (b) Block
map ($64\times64$ 2D and $16\times16$ 3D dimensions, [0.1, 0.3] obstacle fill rate range, [1, 6] number of obstacles range), (c) House atlas ($64\times64$ 2D and $16\times16$ 3D dimensions, [8, 15] minimum room size range, [35, 45] maximum room size range). (Note: Magenta colour is used as goal for all 2D maps as the green goal is difficult to spot.)}
  \label{fig:3maps}  
  \vspace{-2 mm}
  \end{figure}

%%%%%%%%%%%%%%%%%%%%%%%%%%%%%%%%%%%%%%%%%%%%%%%%%%%%%%%%%%%%%
%%%%%%% Supported Algorithms
%%%%%%%%%%%%%%%%%%%%%%%%%%%%%%%%%%%%%%%%%%%%%%%%%%%%%%%%%%%%%
\section{{Supported Path Planning Algorithms}}
\label{sec:supported algorithms}

With the advantage of supporting both classical and machine learning based path planning algorithms, PathBench provides a lightweight framework where development and evaluation of new algorithms can be conducted. Currently supported algorithms are categorized and introduced in the following and additional algorithms can be implemented to PathBench easily.    

\subsection{{Classical Algorithms}}

The classical path planning algorithms implemented into PathBench can be categorized into four different categories. These algorithms include 1) A*, wavefront, Dijkstra planners for graph-based planners; 2) RRT, simple probabilistic roadmap (sPRM) and variations, such as RT, RRT* and RRT-Connect for graph-based category; 3) Bug1 and Bug2~\cite{sensor1997} for sensory-based planning; and 4) potential field algorithm~\cite{pot1992} for numerical optimization category. The numerical optimization approach is also a major component of machine learning methods that are introduced in the learned planning algorithms section that follows. Moreover, additional sampling based algorithms from the Open Motion Planning Library (OMPL) are also added to improve benchmarking capability of PathBench. The algorithms developed inside PathBench support step-by-step planning animation, which is an interesting feature for debugging and educational purposes.

%{\it 1) Graph-Based Planners.} Graph searching path planners utilize graph or occupancy grid as representation of the state space for planning. A*, wavefront, Dijkstra planners are algorithms in the PathBench distribution.

%{\it 2) Sampling-Based Planners.} %Sampling based algorithms generate a simplified configuration space for planning by probing the space with a sampling scheme. 
%Sampling based algorithms including RRT, simple probabilistic roadmap (sPRM) and variations, such as RT, RRT* and RRT-Connect, are developed inside PathBench to support planning animation. Additional sampling based algorithms from the Open Motion Planning Library (OMPL) are also added to improve benchmarking capability of PathBench. 

%{\it 3) Sensory-Based Planners.} Sensory-based planners guarantee convergence by using global information and following locally sensed obstacle boundaries~\cite{Paull_TMech_2013}. PathBench includes Bug1 and Bug2 algorithms as representative for this class of path planning approach~\cite{sensor1997}.  

%{\it 4) Numerical Optimization Planners.} Numerical optimization algorithms describe the path planning problem as a cost function, which is then minimised by using function approximation techniques. Potential field algorithm computes a potential function in which is used to guide path planning ~\cite{pot1992}. Aside from the potential field method included in PathBench, numerical optimization approach is also a major component of machine learning methods that are introduced in the learned planning algorithms section that follows.  

\subsection{{Learned Planning Algorithms}}\label{sec:learnt-alg}

Several machine learning-based path planning algorithms implemented into PathBench explained below.

{\it 1) Value Iteration Networks (VIN)~\cite{tamar2016value}.} VIN is a fully differentiable neural network with an embedded planning module that performs reinforcement learning (RL) for path planning. VIN improves from standard CNN based networks that learned by reactive policies by representing the classical value- iteration planning algorithm in the form of a CNN. The obtained NN model, VIN, provides useful predictions when computing paths.

{\it 2) Motion Planning Networks (MPNet)~\cite{qureshi2019motion}.} MPNet is a neural network based algorithm that uses two neural networks to conduct path planning. The first model, encoder network, embeds the obstacles' point cloud to latent space. The second model, planning network, learns to path plan with the obstacle embedding, start and end position of the robot agent. A hybrid path planning approach combining MPNet and RRT* was developed to improve the success rate of planning in.

{\it 3) Gated Path Planning Networks (GPPN) \cite{lee2018gated}.} GPPN is a deep neural network based path planning approach that improves on VIN. It shows success even on a challenging 3D environment, where the planner is only provided with first-person RGB images.

{\it 4) Online LSTM~\cite{nicola2018lstm}.}
The Online LSTM (long short-term memory) utilizes an LSTM network to determine what action an agent should take next when given the current pose and measurements to achieve the goal. The network takes the start, goal, and current range-bearing measurement as input. This algorithm is greedy, works well on easy maps, but gets in local minimum in complex maps.

{\it 5) CAE-LSTM~\cite{inoue2019robot}.}
The CAE-LSTM adds a convolutional auto-encoder (CAE) component on top of the LSTM network of the Online LSTM \cite{nicola2018lstm} to improve path generation in complex environments and long corridors. The latent variable of the auto-encoder represents a compact representation of the map of the environment. This algorithm works well in complex maps, but deviates from the shortest path.

{\it 6) Bagging LSTM~\cite{wpnconf}.} This algorithm uses an ensemble approach~\cite{dietterich2000ensemble} to get the best out of Online and CAE-LSTM.

{\it 7) Waypoint Planning Networks~(WPN)~\cite{wpnconf}.} WPN integrates three ML-based planning algorithms; Online LSTM, CAE-LSTM, and bagging LSTM; with a waypoint module. WPN makes use of maps and the start and goal positions of the robot as network inputs. Unlike Online LSTM, CAE-LSTM, and bagging LSTM which produce a path, WPN only suggest waypoints towards the goal.

%
% The Online LSTM utilizes an LSTM network to determine what action an agent should take next when given the current location data. 
%
%It is based on \cite{nicola2018lstm}, with some modifications in architecture and logic. %The CAE-LSTM adds a convolutional auto-encoder component on top of the LSTM network in the Online LSTM to improve path generation in complex environments and long corridors. 
%The bagging module combines Online LSTM and CAE-LSTM by using the machine learning ensemble method described in \cite{dietterich2000ensemble}. Finally, the waypoint module utilizes the bagging module as global kernel to propose a series of waypoints for planning. The waypoint module encompasses benefits from all three previously mentioned modules, thus making it the representative WPN algorithm. WPN and its modules are available as separate algorithms in PathBench

\begin{figure}[t]
\centering
    \includegraphics[width=.24\columnwidth]{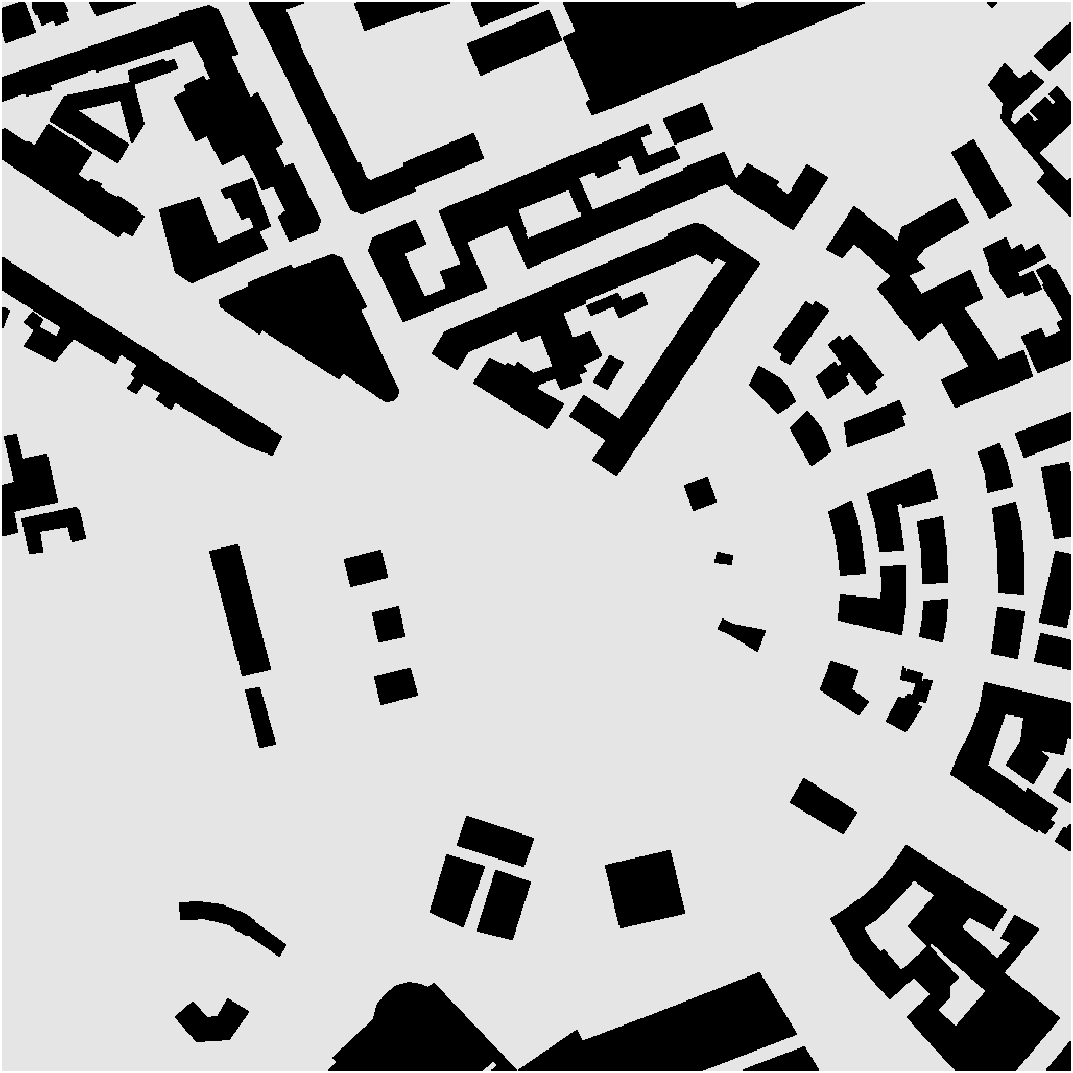}
    \includegraphics[width=.24\columnwidth]{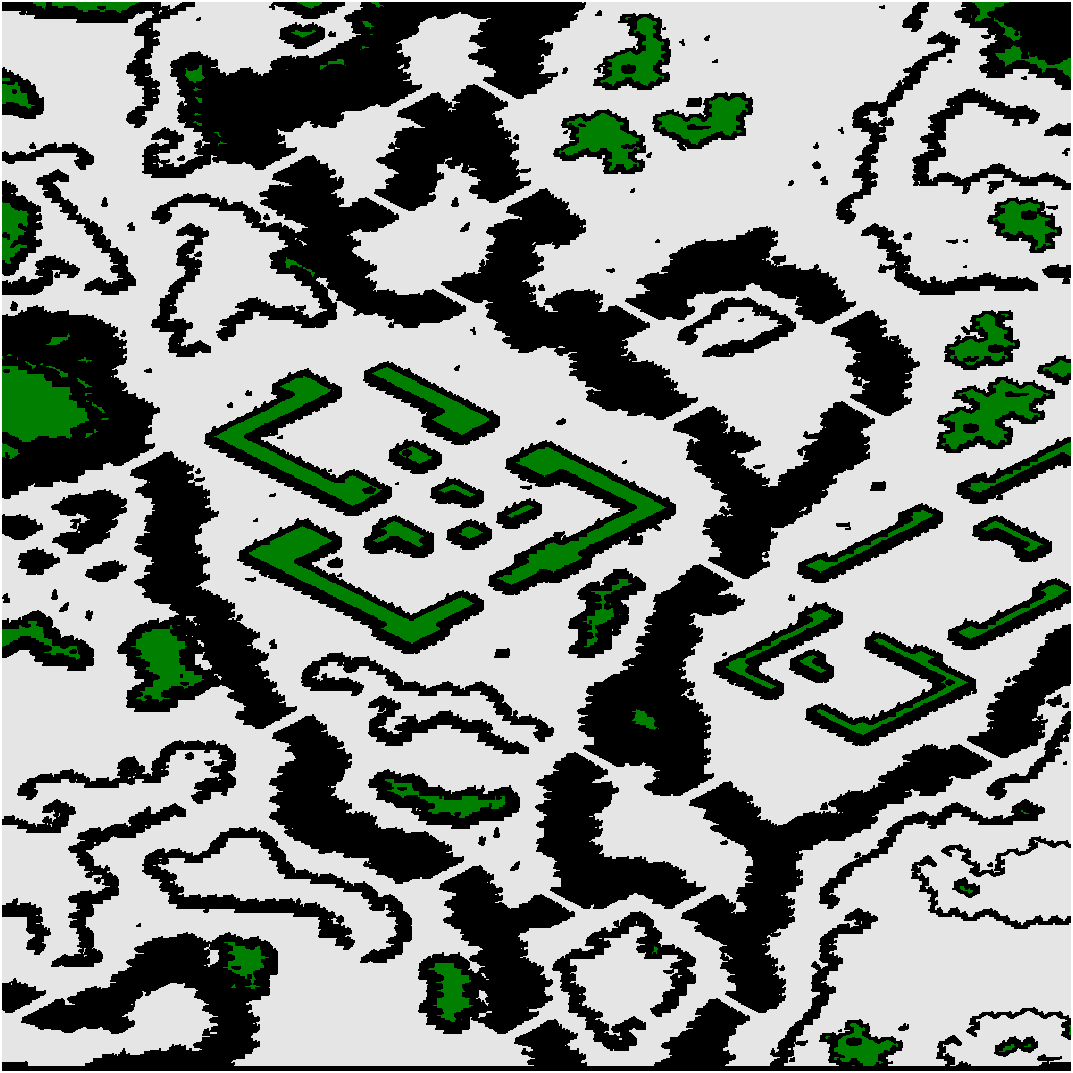}
    \includegraphics[width=.24\columnwidth]{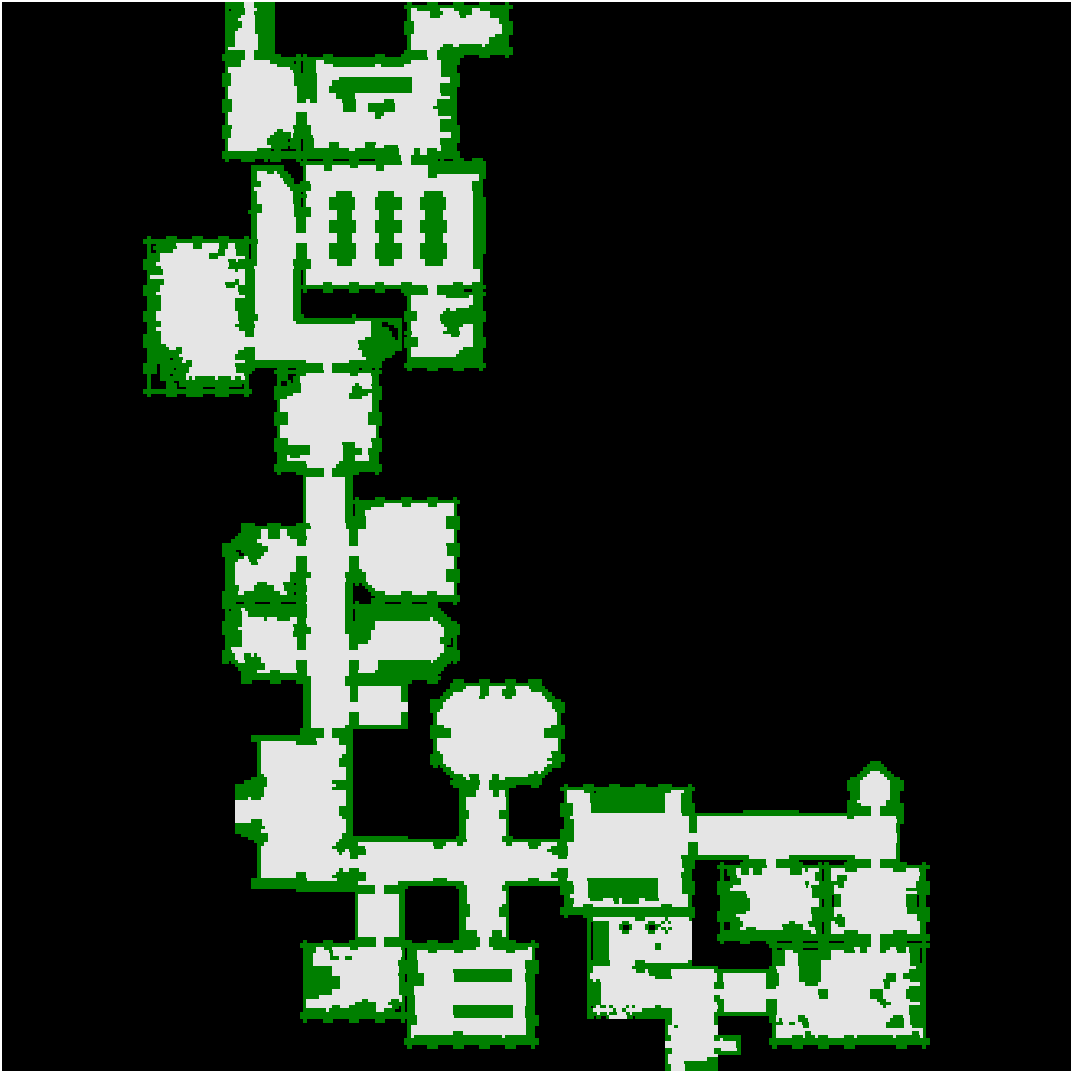}
    %\includegraphics[width=.24\columnwidth]{images/vid4.png}
%\begin{minipage}[b]{0.45\columnwidth}
% \centering
%\includegraphics[width=.96\columnwidth]{images/city1.png} \\
%    \vspace{1mm}  
%\includegraphics[width=.96\columnwidth]{images/vid1.png} \\
%\end{minipage}
% \begin{minipage}[b]{0.45\columnwidth}
%  \centering
%\includegraphics[width=.96\columnwidth]{images/vid2.png}\\
%\vspace{1mm}
%\includegraphics[width=.96\columnwidth]{images/vid4.png} \\
%\end{minipage}
  \vspace{2mm}
\caption{External 
%2D and 3D 
maps can be ported into PathBench for benchmarking with ease. The maps above are from video games and real-world city \cite{sturtevant2012benchmarks}.}
  %\label{fig: wp_vs_a_1}
  \label{fig:extmaps}  
  \vspace{-2 mm}
  \end{figure}

%%%%%%%%%%%%%%%%%%%%%%%%%%%%%%%%%%%%%%%%%%%%%%%%%%%%%%%%%%%%%
%%%%%%% Supported Maps
%%%%%%%%%%%%%%%%%%%%%%%%%%%%%%%%%%%%%%%%%%%%%%%%%%%%%%%%%%%%%
\section{{Supported Maps}}
\label{sec:maps}
The map is the environment of which simulation and benchmarking of algorithms are performed. A Map contains different entities such as the agent, goal and obstacles, and provides a clean interface that defines the movement and interaction between them. Therefore, a map can be extended to support various environments. The following are supported 2D and 3D map types currently in PathBench.

\subsection{Synthetic Maps}
As mentioned previously in the Generator (Sec.~\ref{sec: generator}), four synthetic map types can be created and used inside PathBench. The simplest map type, block maps, contains a random number of randomly sized blocks that act as obstacles. On the other hand, the map type of uniform random fill maps consists of single obstacles placed at random in the maps' free spaces. The third map type, house maps, aims to mimic typical floorplans by placing obstacles in the form of randomly sized and partitioned walls. Lastly, 3D point cloud maps that contain a set of obstacles in an unbounded 3D space can also be generated and used in PathBench. The inclusion of point cloud maps is to facilitate the development and support of algorithms that work exclusively with point clouds, such as MPNet \cite{qureshi2019motion}. Different map types are included in PathBench, so that map-type specific performance of path planning algorithms can be analyzed further (See Fig.~\ref{fig:3maps}.)

\subsection{Real Maps}
Real-world maps can be utilized inside PathBench with the RosMap class. The RosMap extends 2D occupancy grid maps to integrate the gmapping~\cite{gmapping} and other similar 2D SLAM algorithms by converting the SLAM output image into an internal map environment. The RosMap environment has support for live updates, meaning that algorithms can query an updated view by running a SLAM scan. The map uses simple callback functions to
make SLAM update requests and convert movement actions into network messages using the ROS
publisher-subscriber communication system.

\subsection{External Maps}
External maps can be imported into PathBench to diversify the datasets. Houseexpo~\cite{houseexpo} is a large dataset of 2D floor plans built on SUNCG dataset \cite{song2016ssc}. It contains 35,126 2D floor plans that have 252,550 rooms in total and can be used for PathBench benchmarking. In addition, other video game and real-world datasets can also be converted for PathBench use easily. 2D grid world and 3D voxel maps from video games, such as Warcraft III, Dragon Age and Warframe, and real world 2D street maps from OpenStreetMaps geo-spatial database are implemented into PathBench to demonstrate the ease of integrating external datasets \cite{sturtevant2012benchmarks,brewer2018voxels}, see Fig. \ref{fig:extmaps}. Benchmarking results on external maps are shown in Sec. \ref{sec:result}.

%%%%%%%%%%%%%%%%%%%%%%%%%%%%%%%%%%%%%%%%%%%%%%%%%%%%%%%%%%%%%
%%%%%%% Performance Metrics
%%%%%%%%%%%%%%%%%%%%%%%%%%%%%%%%%%%%%%%%%%%%%%%%%%%%%%%%%%%%%
\section{{Performance Metrics}}
\label{sec:performance metrics}

In order to evaluate and benchmark performance of various algorithms inside PathBench, several metrics are chosen, including success rate, path length, distance left to goal when failed, time, path deviation, search space, memory consumption, obstacle clearance and smoothness of trajectory. Algorithm selection can be aided by evaluating the benchmarked results of task-specific metrics. The following outlines the metrics and rationales behind their selection. 

{\it 1) Success Rate (\%).} The rate of success of finding a path, from start to goal, demonstrates the reliability of the algorithm.  

{\it 2) Path Length (metres).} The total distance taken to reach the goal showcases the efficiency of the path generated.   

{\it 3) Distance Left To Goal (metres).} The Euclidean distance left
from the agent to the goal, in case of a algorithm failure. This shows the extent of the planning failure.  

{\it 4) Time (seconds).} The total time taken to reach the goal. Time required for planning is an important factor for real life robotics applications. 

{\it 5) Path Deviation (\%).} The path length difference when compared to the shortest possible path, generated by A*. Allows comparison to an "optimal" planner.  

{\it 6) Search Space (\%).} The amount of space that was explored and used to find the final path to the goal. 

{\it 7) Maximum Memory Consumption (MB).} The maximum amount of memory used during a path generation session. Memory usage could be a limiting factor for various robotics settings, thus being a relevant benchmarking metric. 

{\it 8) Obstacle clearance (metres).} Obstacle clearance provides the mean distance of the agent from obstacles during traversal.

{\it 9) Smoothness of trajectory (degrees).} The average angle change between consecutive segments of paths shows how drastic and sudden the agent's movement changes could be.

Other than the metrics above, additional metrics can be implemented into PathBench if required. Nowak {\it et al.} provide potential metrics that could be added, including orientation error, number of collisions,  number of narrow passages traversed and number of parameters to tune \cite{nowak2010}.

%%%%%%%%%%%%%%%%%%%%%%%%%%%%%%%%%%%%%%%%%%%%%%%%%%%%%%%%%%%%%
%%%%%%% Experimental Results
%%%%%%%%%%%%%%%%%%%%%%%%%%%%%%%%%%%%%%%%%%%%%%%%%%%%%%%%%%%%%
\section{Experimental Results}
\label{sec:result}

% In this section, experimental results are presented. The experiments include: path planning in simulated and real-world occupancy grid maps, and path planning and navigation using a real robot in real-world office environment. The training of the algorithm is performed using simulated maps only.

% \subsection{Simulated Environment}
% In this section, first the simulated dataset is briefly explained, then the metrics to evaluate the algorithm are described.

% \paragraph{Dataset.}
% For training, three types of synthetic map of size 64 $\times$ 64 pixels were procedurally generated: uniform random fill map, block map, and house map. The map generation process with sample maps are explained in the Supplementary Materials. %Sample maps are shown in Figure \ref{fig: eval_generated maps}. %The analysis of the training datasets can be found in Table \ref{tab: eval_maps}.
% Fig.~\ref{fig: Way runs comp} shows samples of these maps. In these maps, start and goals point are chosen randomly. Evaluations are done over maps that has never been seen by the algorithm. 

% \paragraph{Evaluation Metrics.} To evaluate the generated maps four metrics are used: success rate, trajectory length, distance left to goal when failed, and session search space. To demonstrate why such an architecture was chosen for WPN, we compare the proposed algorithm not only with A*, but also with other modules described in the paper, mainly \emph{Online LSTM}, \emph{CAE-LSTM} and \emph{bagging module}.

In this section, several experiments using PathBench, with classical and learned planners on different maps are presented. %The algorithms are evaluated on 2D and 3D synthetic maps of varying sizes that are generated inside PathBench. In addition, video game and street maps from external datasets \cite{sturtevant2012benchmarks} are used for further algorithm benchmarking. Finally, algorithms are tested in ROS and Gazebo with PathBench's ROS extension to highlight its ability to integrate with real world robotics applications. 

\subsection{Algorithmic Benchmarking}
Classical and learned algorithms, currently supported by PathBench, are benchmarked inside PathBench with different types of maps. All results are produced by PathBench on Ubuntu 18.04 with Intel Core i5-6200U CPU and an Nvidia GeForce 940MX.  
For training of the learned algorithms, 
%WPN, VIN, MPNet and GPPN, 
three types of synthetic map of size $64 \times 64$ pixels were procedurally generated: uniform random fill map, block map, and house map. 
%
%The map generation process with sample maps are explained in the Supplementary Materials. 
%
%Sample maps are shown in Figure \ref{fig: eval_generated maps}. %The analysis of the training datasets can be found in Table \ref{tab: eval_maps}.
Fig.~\ref{fig:3maps} shows samples of these maps. In these maps, start and goal points are chosen randomly. Evaluations are done on maps that have never been seen by the algorithm. 
%
%\textbf{Evaluation Metrics.} 
%To evaluate the generated maps, nine metrics are used: success rate, path length, distance left to %goal when failed, time, path deviation, search space, obstacle clearnace, smoothness of trajectory %and memory consumption.

%\begin{figure}
%  \hfill
%  \begin{minipage}[b]{0.16\textwidth}
%  \centering
%    \vspace{-2 mm}
%    \includegraphics[width=\linewidth]{images/m1.png}
%  \end{minipage}
%  \hfill
%  \begin{minipage}[b]{0.16\textwidth}
%  \centering
%    \vspace{-2 mm}
%    \includegraphics[width=\linewidth]{images/m2.png}
%  \end{minipage}
%  \hfill
%  \begin{minipage}[b]{0.16\textwidth}
%  \centering
%    \vspace{-2 mm}
%    \includegraphics[width=\linewidth]{images/m3.png}
%  \end{minipage}
%  \hfill
%  \begin{minipage}[b]{0.16\textwidth}
%  \centering
%    \vspace{-2 mm}
%    \includegraphics[width=\linewidth]{images/m4.png}
%  \end{minipage}
%  \hfill
%  \begin{minipage}[b]{0.16\textwidth}
%  \centering
%    \vspace{-2 mm}
%    \includegraphics[width=\linewidth]{images/m5.png}
%  \end{minipage}
%  \hfill
%  \begin{minipage}[b]{0.16\textwidth}
%  \centering
%    \vspace{-2 mm}
%    \includegraphics[width=\linewidth]{images/m6.png}
%  \end{minipage}
%  \hfill  
%\caption{Synthetic maps used to generate results in Table II.}
%  \label{fig:synthetic_maps}  
%  \end{figure}

\begin{table}[t]
    \centering
    \caption{Results of classical algorithms: On 2D 64$\times$64 PathBench built in maps, 512$\times$512 city maps, and video game maps with 800 to 1200 cells in dimension. The failed cases occur when there is no valid path towards the given goal.}
    \vspace{1mm}
    \begin{tabular}{|c|c|c|c|c|c|}
       \hline
       %\textbf{\scriptsize planner} & \makecell {\textbf{\scriptsize success} \\ \textbf{\scriptsize %rate} } & \makecell {\textbf{\scriptsize path len.} \\ \textbf{\scriptsize }{\scriptsize} } & %\makecell {\textbf{\scriptsize distance left} \\ \scriptsize{(when failed)} } &  %\makecell{\textbf{\scriptsize time} \\ (sec)} & \makecell {\textbf{\scriptsize path} \\ %\textbf{\scriptsize deviation} }  \\
    \makecell {\textbf{\scriptsize map} \\ \textbf{\scriptsize type}} & \textbf{\scriptsize planner} & \makecell {\textbf{\scriptsize path} \\ \textbf{\scriptsize dev.}\scriptsize{(\%)}}  & \makecell {\textbf{\scriptsize distance left} \\ (if failed,m)} &\makecell{\textbf{\scriptsize time} \\ (sec)} & \makecell {\textbf{\scriptsize success} \\ \textbf{\scriptsize rate} \scriptsize{(\%)} }\\ 
       \hline
      \parbox[t]{2mm}{\multirow{7}{*}{\rotatebox[origin=c]{90}{PathBench Maps}}} 
      &{\scriptsize A* \cite{duchovn2014path}}  & 0.00 & 0.00 & 0.103 & 100.0  \\%& 10.5\% \\
       \cline{2-6}
       &{\scriptsize Wavefront \cite{luo2014effective}}  & 0.34 & 0.00 & 0.334 & 100.0 \\%& -\% \\
       \cline{2-6}
       &{\scriptsize Dijkstra \cite{choset2005principles}}  & 0.00 & 0.00 & 0.578  & 100.0 \\%& 42.7 \\
       \cline{2-6}
       &{\scriptsize SPRM \cite{kavraki1994probabilistic}}   & 30.24 & 1.87 & 0.596 & 95.0 \\%& 42.7 \\
       \cline{2-6}
       &{\scriptsize RRT~\cite{lavalle1998rapidly}} & 13.11 & 1.83 & 7.334 & 97.6 \\%& - \\
       \cline{2-6}
       &{\scriptsize RRT* \cite{rrtstar}} & 6.29 & 1.46 & 9.412  & 94.3 \\%& - \\
       \cline{2-6}
       &{\scriptsize RRT-Connect~\cite{rrtconnect}} & 17.77 & 0.18 & 0.137 & 99.5  \\%& - \\
       \hline
       \hline
       \parbox[t]{2mm}{\multirow{7}{*}{\rotatebox[origin=c]{90}{City Maps}}} 
       &{\scriptsize A*}            & 0.00 & 0.00 & 3.816 & 100.0 \\%& 10.5\% \\
       \cline{2-6}
       &{\scriptsize Wavefront}     & 1.08  & 0.00 &  8.468 & 100.0\\%& -\% \\
       \cline{2-6}
       &{\scriptsize Dijkstra}   & 0.00 & 0.00 & 9.928 & 100.0 \\%& 42.7 \\
       \cline{2-6}
       &{\scriptsize SPRM}   & 123.64 & 13.68 & 5.377 & 93.6  \\%& 42.7 \\
       \cline{2-6}
       &{\scriptsize RRT} & 54.98 & 3.43 & 39.248 & 95.7 \\%& - \\
       \cline{2-6}
       &{\scriptsize RRT* } & 32.71 & 4.11 & 43.464  & 95.1 \\%& - \\
       \cline{2-6}
       &{\scriptsize RRT-Connect} & 65.08 & 5.35 & 3.489 & 96.6 \\%& - \\
       \hline
       \hline
       \parbox[t]{2mm}{\multirow{7}{*}{\rotatebox[origin=c]{90}{Video Game Maps}}} 
      &{\scriptsize A*}               & 0.00 & 0.00 & 31.567 & 100.0 \\%& 10.5\% \\
       \cline{2-6}
       &{\scriptsize Wavefront}     & 1.65 & 0.00 & 43.517 & 100.0 \\%& -\% \\
       \cline{2-6}
       &{\scriptsize Dijkstra}   & 0.00 & 0.00 & 42.366 & 100.0 \\%& 42.7 \\
       \cline{2-6}
       &{\scriptsize SPRM}   & 287.21 & 0.00 & 44.498 & 100.0 \\%& 42.7 \\
       \cline{2-6}
       &{\scriptsize RRT} & 128.90 & 64.38 & 64.881 & 42.6 \\%& - \\
       \cline{2-6}
       &{\scriptsize RRT* } & 104.30 & 56.21 & 68.971  & 36.7 \\%& - \\
       \cline{2-6}
       &{\scriptsize RRT-Connect} & 112.90 & 25.92 & 30.84 & 95.30 \\%& - \\
       \hline
%       {\scriptsize Iterative  \cite{qureshi2018motion}} & \% &  &  &  &- \\
%       \hline
    \end{tabular}
    \label{tab:classicresult}
        \vspace{-2 mm}
\end{table}

\begin{table}[t]
    \centering
    \caption{Results of learned algorithms: On the same 2D 64$\times$64 PathBench built in maps, 512$\times$512 city maps, and video game maps from Table~\ref{tab:classicresult}.}
    \vspace{1mm}
    \begin{tabular}{|c|c|c|c|c|c|}
       \hline
       %\textbf{\scriptsize planner} & \makecell {\textbf{\scriptsize success} \\ \textbf{\scriptsize %rate} } & \makecell {\textbf{\scriptsize path len.} \\ \textbf{\scriptsize }{\scriptsize} } & %\makecell {\textbf{\scriptsize distance left} \\ \scriptsize{(when failed)} } &  %\makecell{\textbf{\scriptsize time} \\ (sec)} & \makecell {\textbf{\scriptsize path} \\ %\textbf{\scriptsize deviation} }  \\
    \makecell {\textbf{\scriptsize map} \\ \textbf{\scriptsize type}} & \textbf{\scriptsize planner} & \makecell {\textbf{\scriptsize path } \\ \textbf{\scriptsize dev.}\scriptsize{(\%)}}  & \makecell {\textbf{\scriptsize distance left} \\ (if failed,m)} &\makecell{\textbf{\scriptsize time} \\ (sec)} & \makecell {\textbf{\scriptsize success} \\ \textbf{\scriptsize rate}\scriptsize{(\%)}}\\ 
       \hline
      \parbox[t]{2mm}{\multirow{7}{*}{\rotatebox[origin=c]{90}{PathBench Maps}}} 
      &{\scriptsize VIN \cite{tamar2016value}}      &76.31 & 21.80 & 0.583 & 28.7   \\%&  \\
       \cline{2-6}
       &{\scriptsize MPNet \cite{qureshi2019motion} } & 28.17 & 32.60 &  0.988 & 21.4  \\%& - \\
       \cline{2-6}
       &{\scriptsize GPPN \cite{lee2018gated}} & 5.81  & 26.11 & 5.813 & 35.7  \\%& - \\
       \cline{2-6}
       &{\scriptsize Online LSTM~\cite{nicola2018lstm} }   & 0.45  & 7.65 & 0.172 & 67.6\\%& 42.7 \\
       \cline{2-6}
       &{\scriptsize CAE-LSTM~\cite{inoue2019robot}}   & 0.50 & 9.27 & 0.216 & 61.3\\%& 42.7 \\
       \cline{2-6}
       &{\scriptsize Bagging LSTM \cite{wpnconf}} & 1.63 & 1.41 & 1.052 & 92.3 \\%& - \\
       \cline{2-6}
       &{\scriptsize WPN \cite{wpnconf}}             & 1.86 & 0.00 & 0.617 & 100.0 \\%& -\% \\
       \hline
       \hline
       \parbox[t]{2mm}{\multirow{7}{*}{\rotatebox[origin=c]{90}{City Maps}}} 
      &{\scriptsize VIN}               & 100.00 & 184.21 & 32.612 & 0.0 \\%& 10.5\% \\
       \cline{2-6}
       &{\scriptsize GPPN } & 100.00 & 184.21 &  57.633 & 0.0 \\%& - \\     
       \cline{2-6}
       &{\scriptsize Online LSTM}   & 1.05 & 87.51 & 3.328 & 16.7\\%& 42.7 \\
       \cline{2-6}
       &{\scriptsize CAE-LSTM}   & 4.49 & 91.20 & 4.241 & 10.0\\%& 42.7 \\
       \cline{2-6}
       &{\scriptsize Bagging LSTM} & 31.67 & 45.06 & 20.268 & 43.3  \\%& - \\
       \cline{2-6}
       &{\scriptsize WPN}             & 8.44 & 0.00 & 13.816  & 100.0 \\%& -\% \\
       \hline
       \hline
       \parbox[t]{2mm}{\multirow{7}{*}{\rotatebox[origin=c]{90}{Video Game}}} 
      &{\scriptsize VIN}               & 100.00 & 276.30 & 42.19 & 0.0 \\%& 10.5\% \\
       \cline{2-6}
       &{\scriptsize GPPN } & 100.00 & 276.30 & 65.877 & 0.0 \\%& - \\
       \cline{2-6}
       &{\scriptsize Online LSTM}   & 0.00 & 217.10 & 16.380 & 8.3 \\%& 42.7 \\
       \cline{2-6}
       &{\scriptsize CAE-LSTM}   & 3.05 & 199.60 & 27.330 & 5.3\\%& 42.7 \\
       \cline{2-6}
       &{\scriptsize Bagging LSTM } & 0.41 & 155.89 & 123.901 & 20.6  \\%& - \\
       \cline{2-6}
       &{\scriptsize WPN}             & 10.55 & 0.00  & 110.307  & 100.0 \\%& -\% \\
       \hline
%       {\scriptsize Iterative  \cite{qureshi2018motion}} & \% &  &  &  &- \\
%       \hline
    \end{tabular}
    \label{tab:learnedresult}
        \vspace{-2 mm}
\end{table}

\begin{figure}[t]
    \includegraphics[width=1.0\columnwidth]{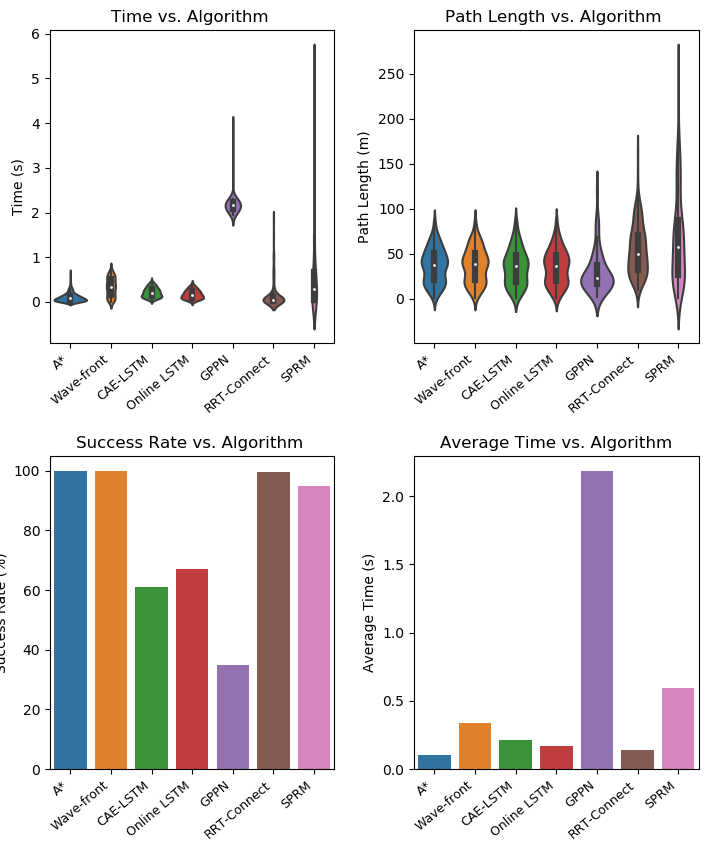}
  \vspace{-4mm}
\caption{Graphical analysis of 2D benchmarking for classical and learned algorithms.}
  %\label{fig: wp_vs_a_1}
  \label{fig:barvio}  
  \vspace{-1 mm}
  \end{figure}

\subsubsection{2D Synthetic Maps: Simple Analysis}
To demonstrate the benchmarking ability of PathBench and its support for the machine learning algorithms, 
%WPN, Online LSTM, CAE-LSTM, bagging module, VIN, GPPN and MPNet planners
all the algorithms described in Sec.~\ref{sec:learnt-alg} are analyzed against classical path planning algorithms in $64 \times 64$ 2D PathBench maps. One thousand maps of each of the three types of PathBench maps were used. Table~\ref{tab:classicresult} and Table~\ref{tab:learnedresult} present detailed comparative results for simple analysis of 3000 2D PathBench maps. Fig.~\ref{fig:barvio} displays some of the key results in bar and violin plots.

\subsubsection{2D External Maps: Complex Analysis}
Both classical and learned algorithms were also benchmarked using the analyzer's complex analysis tool, in order to demonstrate the framework's ability to evaluate algorithm performance on specific map types. The analysis was performed on $n=30$ external city maps from OpenStreetMaps' geo-spatial database \cite{sturtevant2012benchmarks}, with 10 random samples collected for averaging of results on each $512 \times 512$ map. Thirty video game maps with height and width varying from 800 to 1200 cells were benchmarked in a similar manner. Results of benchmarking on video game and city maps are also listed in Table~\ref{tab:classicresult} and Table~\ref{tab:learnedresult}. The use of external environments in this experiment demonstrates the capability of PathBench to incorporate additional datasets. 

\subsubsection{3D Maps: Simple Analysis}
To demonstrate PathBench's support for 3D path planning, analysis of path planning algorithms on 3D $28 \times 28 \times 28$ PathBench maps was conducted. The benchmarking results that averaged algorithm performance on 1000 maps of each PathBench map type, uniform random fill map, block map, and house map, is shown in Table~\ref{tab:3dresults}.

By looking at the results, we can quickly assess some strengths and weaknesses of each planning approach. For example, the three graph-based algorithms all find a solution 100\% of the time, provided one exists. RRT* paths are always shorter than those of RRT as expected, and RRT-Connect has much higher success rate than any other sampling-based method, while being considerably faster. The number of samples taken is a parameter that can be modified easily to configure sampling-based algorithms' behaviour. Although A* can generate the shortest path length in both 2D and 3D planning scenarios, RRT-Connect is capable of planning at a significantly faster time in 3D environments. Machine learning algorithms, on the other hand, experience lower success rates for all map types. VIN and GPPN have shown to not scale well with the increase in map size and could not successfully provide any paths in the city and video game datasets. WPN is an exception with the ability to plan at 100\% success rate for all map types used. However, machine learning algorithms' path planning times in general are higher when compared to classical approaches, especially as the map size increases. MPNet was only tested on PathBench maps, due to constraint of the implementation used. The open source version of the network only allowed encoding of a limited number of obstacles. Performing this kind of simple and rapid analysis is trivial in PathBench. Furthermore, we can notice that the algorithms maintain their same behaviour across different environments.

%%%%%%%%%%%%%%%%%%%%%%%%%%%%%%%
%%%%%%%%% Table 3D results
%%%%%%%%%%%%%%%%%%%%%%%%%%%%%%%
%%%%%%%%%%%%%%%%%%%%%%%%%%%%%%%

\begin{table}[t]
    \centering
    \caption{Results of classical algorithms on 3D 28$\times$28$\times$28 PathBench built in maps.}
    \vspace{1mm}
    \begin{tabular}{|c|c|c|c|c|c|}
       \hline
       \textbf{\scriptsize planner} & \makecell {\textbf{\scriptsize success} \\ \textbf{\scriptsize rate}{\scriptsize (\%)} } & \makecell {\textbf{\scriptsize path } \\ {\scriptsize \textbf{len.}(m)} } & \makecell {\textbf{\scriptsize path} \\ \textbf{\scriptsize dev. {\scriptsize(\%)}} }&  \makecell{\textbf{\scriptsize time} \\ (sec)} & \makecell {\textbf{\scriptsize path smooth-} \\ \textbf{\scriptsize ness}{\scriptsize (deg.)}}   \\
       \hline
       {\scriptsize A*}               & 100.0 & 20.69 & 0.00 & 0.475 & 0.28 \\
       \hline
       {\scriptsize Wavefront}              & 100.0 & 21.27  &  0.51 & 6.118 & 0.11 \\
       \hline
       {\scriptsize Dijkstra}   & 100.0 & 20.69 & 0.00 & 8.453 & 0.13 \\
       \hline
       {\scriptsize SPRM}   & 100.0 & 36.87 & 16.18 & 0.248 & 0.37 \\
       \hline
       {\scriptsize RRT-Connect} & 99.7 & 38.22 & 17.56 & 0.097 & 0.41 \\
       \hline
%       {\scriptsize Iterative  \cite{qureshi2018motion}} & \% &  &  &  &- \\
%       \hline
    \end{tabular}
    \label{tab:3dresults}
        \vspace{-2 mm}
\end{table}

\subsection{Real-world Robot Interfacing}
PathBench has the capability to natively interface with ROS and Gazebo to allow for seamless path planning for simulated and real world robotic applications. PathBench is able to visualize and plan for both fixed map environements and exploration environments. The planning is done in PathBench in real-time, and the control commands are sent to ROS to guide the robot. 
% The robot environment is simulated in Gazebo as a house style map, and the can also be visualized through Rviz, all launched from within PathBench. A sample experiment can be seen in Fig.~\ref{fig:gazebo_ros}. See the supplementary video and GitHub for more visualizations. 
% 
We can also demonstrate the live map capabilities of PathBench, using an algorithm with exploration capabilities. We use WPN-view for our testing. The robot is able to plan into known space, and also plan into the unknown environment, while the PathBench map is updated as it explores. This can also be seen in the supplementary video and GitHub, where the exploration is demonstrated.

\vspace{-1 mm}
\section{Conclusion}
\label{sec:conclusion}
PathBench presents a significant advantage in terms of developing and evaluating classical and learned motion planning algorithms, by providing development environment and benchmarking tools with standard and customizable metrics. PathBench has been demonstrated across a wide range of algorithms and datasets.

In the future, PathBench will be extended to allow benchmarking and training of additional learning-based algorithms, along with support for higher dimensional planning with constraints. %, to eliminate the effort required to obtain such results.
%Moreover, we will build a large dataset of 2D and 3D maps, extracted from datasets such as %InteriorNet.

%===============================================================================

% The maximum paper length is 8 pages excluding references and acknowledgements, and 10 pages including references and acknowledgements

% The acknowledgments are automatically included only in the final version of the paper.

%===============================================================================


\section*{Acknowledgment}
This work was partially funded by  DRDC-IDEaS (CPCA-0126) and EPSRC (EP/P010040/1). We acknowledge the technical contributions made by 
Judicael E Clair, 
Danqing Hu, 
Radina Milenkova, 
Zeena Patel, 
Abel Shields, and 
John Yao to the programming of the project and producing of
Fig.~\ref{fig: sim}-e-f, 
Fig.~\ref{fig: simulator3d}, and Fig.~\ref{fig:3maps}'s second row. Notable programming contributions made include the added support of PathBench's rendering and support of 3D path planning environments, infrastructural changes for efficiency and a renewed ROS interface.    

\bibliographystyle{IEEEtran}
\bibliography{main}

\end{document}